\documentclass{bmvc2k}

\title{ViewNeRF: Unsupervised Viewpoint Estimation Using Category-Level Neural Radiance Fields}

\addauthor{Octave Mariotti}{omariott@ed.ac.uk}{1}
\addauthor{Oisin Mac Aodha}{oisin.macaodha@ed.ac.uk}{1}
\addauthor{Hakan Bilen}{hbilen@ed.ac.uk}{1}

\addinstitution{
 School of informatics\\
 University of Edinburgh\\
 Edinburgh, UK
}
\runninghead{Mariotti, Mac Aodha, Bilen}{ViewNeRF}

\def\eg{\emph{e.g}\bmvaOneDot}

\def\etc{\emph{etc}\bmvaOneDot}
\def\ie{\emph{i.e}\bmvaOneDot}

\usepackage{textcomp, gensymb}
\usepackage{amssymb}
\usepackage[ruled,vlined]{algorithm2e}
\usepackage[capitalize]{cleveref}
\usepackage{booktabs}
\SetKwInput{KwInput}{Input}                
\SetKwInput{KwOutput}{Output}              

\usepackage{multirow}
\usepackage{rotating}
\usepackage{subcaption}
\usepackage[euler]{textgreek}
\begin{document}

\maketitle

\begin{abstract}
We introduce ViewNeRF, a Neural Radiance Field-based viewpoint estimation method that learns to predict category-level viewpoints directly from images during training. While NeRF is usually trained with ground-truth camera poses, multiple extensions have been proposed to reduce the need for this expensive supervision. Nonetheless, most of these methods still struggle in complex settings with large camera movements, and are restricted to single scenes, \ie they cannot be trained on a collection of scenes depicting the same object category. To address these issues, our method uses an analysis by synthesis approach, combining a conditional NeRF with a viewpoint predictor and a scene encoder in order to produce self-supervised reconstructions for whole object categories. Rather than focusing on high fidelity reconstruction, we target efficient and accurate viewpoint prediction in complex scenarios, \eg 360\degree~rotation on real data. Our model shows competitive results on synthetic and real datasets, both for single scenes and multi-instance collections.
\end{abstract}


\section{Introduction}
\label{sec:intro}
Understanding the 3D world from images is a longstanding goal in computer vision.
Recovering camera viewpoint (\ie pose) is a key step in establishing the correspondence between 2D images and the 3D world. 
Camera viewpoint is typically defined relative to a coordinate system in a single 3D scene or to a canonical frame in an object category.
The former case often involves analysis from multiple views of the same scene (\eg Structure from Motion (SfM)~\cite{longuet1981computer}) where recovered viewpoints are tied to a single scene.
The latter case aims to estimate viewpoint relative to unseen object instances from a specific category (\eg \cite{lowe1987three,huttenlocher1987object}).
It is often more challenging, as it requires invariance to not only view-dependent changes but also to shape and texture variations among instances of an object category. 
In this paper, we focus on the latter case, \emph{category-level viewpoint estimation}.

A notable difficulty in learning to estimate viewpoint is obtaining reliable ground truth poses.
This process is typically complex, requiring either time-consuming manual annotations or calibrated laboratory setups, which limits the applicability of supervised learning.
To this end, multiple approaches have been proposed to learn category-level poses from image collections \emph{without} ground-truth supervision~\cite{tulsiani2018multi, insafutdinov2018unsupervised, mustikovela2020self, mariotti2021viewnet}. 
These works broadly use an analysis-by-synthesis approach by disentangling viewpoint and 3D appearance from images and render objects from new viewpoints, hence enabling the learning of object poses without external pose supervision. 
As the supervision from the reconstruction relies on projection of the 3D scene, it is essential that these methods are able to model 3D information as accurately as possible. 
However, prior work either builds on simplistic pseudo-rendering methods through neural decoders~\cite{nguyen2019hologan, mariotti2020semi, mustikovela2020self} that do not preserve the 3D geometry in rendering, or on classical 3D representations such as voxels~\cite{tulsiani2018multi, mariotti2021viewnet} or point clouds \cite{insafutdinov2018unsupervised} that fail to produce high-quality reconstructions.

Recently, neural radiance fields (NeRF)~\cite{mildenhall2020nerf} have achieved unprecedented quality in the 3D reconstruction and rendering of scenes. 
Deviating from the traditional geometrically-explicit representations, NeRF belongs to the family of implicit 3D representations~\cite{park2019deepsdf, mescheder2019occupancy, chen2019learning, sitzmann2019scene, lombardi2019neural}, encoding 3D data in the weights of a neural network which allows them to operate at continuous 3D coordinates and hence at high resolution. 
In addition, their viewpoint-dependent rendering enables the modeling of material properties such as reflections.
Despite the recent progress~\cite{liu2020neural,barron2021mip, martin2021nerf,sitzmann2021light, yu2021plenoxels, yu2021plenoctrees}, they still typically require accurate camera poses during training, and are also limited to modeling a single scene at a time. 
This restricts their application to controlled settings, with labeled poses, or to scenes where many high-quality camera poses can be obtained via SfM.
Recent works, which aim to alleviate these problems, can either operate only on simple forward-facing scenes without accurate camera pose initialization or only work on a single scene or object instance~\cite{wang2021nerf--, jeong2021self-cal, lin2021barf, meng2021gnerf}. 
Those that model multiple instances, require ground-truth camera poses and/or expensive test-time optimization in order to synthesize novel views~\cite{yu2021pixelnerf,jang2021codenerf} (see~\cref{tab:models_table}).
Hence, these models cannot be trivially used to learn category-level poses without training  pose supervision. 

Motivated by these limitations, we propose an analysis-by-synthesis approach that leverages the powerful 3D modeling ability of NeRFs for unsupervised category-level viewpoint estimation.
Specifically, our model, \emph{ViewNeRF} (i)~simultaneously learns to estimate shape, appearance, and pose of object instances, enables single-shot pose prediction on unseen views more efficiently than the gradient descent-based pose estimation used in methods like~\cite{yen2020inerf, wang2021nerf--,jeong2021self-cal, lin2021barf, jang2021codenerf}, 
(ii)~works on multiple instances of an object category via a conditional NeRF model and extends the classic single-scene setting used in pose-free NeRFs~\cite{meng2021gnerf}, 
and (iii)~obtains significantly more accurate pose predictions compared to  existing unsupervised method~\cite{mariotti2021viewnet} across multiple benchmarks.

\section{Related work}
\label{sec:relwork}

\noindent{\bf Unsupervised viewpoint estimation.} 
Despite the large body of supervised methods~\cite{rad2017bb8, kehl2017ssd, tulsiani2015viewpoints, choy20163d}, viewpoint estimation is still a challenging task due to the cost of building large labeled datasets. 
Hence a growing number of methods attempts to limit the amount of supervision needed at training time.
SfM approaches such as COLMAP~\cite{schonberger2016structure} use multi-view geometry to infer camera poses using only images, but are limited to single scenes, require many views, and are thus unsuitable for estimating poses across category-centric datasets. 
Recently several deep learning pose-estimation methods have been proposed that utilize various amounts of pose supervision including semi-supervised~\cite{mariotti2020semi, wang2021neural}, few/zero-shot learning~\cite{xiao2020few, banani2020novel, goodwin2022zero}, and unsupervised methods~\cite{tulsiani2018multi, insafutdinov2018unsupervised, mariotti2021viewnet, mustikovela2020self}. 
Most related to us,  unsupervised methods typically learn to disentangle category-level pose and appearance using an analysis-by-synthesis pipeline. 
ViewNet~\cite{mariotti2021viewnet} generates a voxel-based reconstruction of a specific object instance and renders it from the predicted viewpoint. 
This approach is limited by the spatial resolution of the voxel grid, fails to reconstruct fine details and viewpoint dependent illumination effects. 
SSV~\cite{mustikovela2020self} adopts a generative approach, using 3D latent feature maps to represent the scene. 
However, it does not enforce geometric consistency between decoded viewpoints, resulting in noisy viewpoint estimation due to its decoder's flexibility and ability to overfit to geometrically implausible poses.

\begin{table}[t]
    \centering
        \resizebox{0.7\linewidth}{!}{
    \begin{tabular}{l|c|c|c|c}
                                                 & \shortstack{Pose-free\\360\degree~training} & \shortstack{Real data\\360\degree~training} & \shortstack{One shot pose\\on new views}  & \shortstack{Multiple\\scenes}\\ \hline
        NeRF~\cite{yen2020inerf}                 &               &               &                       &            \\
        INeRF~\cite{yen2020inerf}                &               &               &                       &            \\
        NeRF--~\cite{wang2021nerf--}$^{\dagger}$ &               &               &                       &            \\
        BARF~\cite{lin2021barf}                  &               &               &                       &            \\
        SCNeRF~\cite{jeong2021self-cal}          &               & \checkmark    &                       &            \\ 
        GaRF~\cite{chng2022garf}$^{\dagger}$     &               &               &                       &            \\ 
        GNeRF~\cite{meng2021gnerf}               & \checkmark    &               & \checkmark$^{\ddag}$  &            \\
        CodeNeRF~\cite{jang2021codenerf}         &               &               &                       & \checkmark \\
        ViewNeRF (Ours)                                     & \checkmark    & \checkmark    & \checkmark            & \checkmark \\
        
    \end{tabular}
    }
    \vspace{10pt}
    \caption{Comparison of pose-free NeRF methods on 360\degree scenes. Most of these approaches require ground-truth poses or initial estimates.
    $^{\dagger}$Untested on 360\degree scenes. $^{\ddag}$Contains a pose estimator but it is not used during evaluation.
    }
    \label{tab:models_table}
    \vspace{-10pt}
\end{table}

\noindent{\bf Neural Radiance Fields (NeRFs).} 
3D data is traditionally represented using meshes~\cite{kanazawa2018learning, kato2018neural}, voxels~\cite{yan2016perspective, tulsiani2017multi, tulsiani2018multi, sitzmann2019deepvoxels}, or point clouds~\cite{insafutdinov2018unsupervised}. 
Recently, implicit representations~\cite{park2019deepsdf, mescheder2019occupancy, chen2019learning, sitzmann2019scene, lombardi2019neural} have emerged as an effective tool in 3D modeling. 
They represent 3D data implicitly in the parameters of a fully connected neural network that takes 3D coordinates as input and predicts properties such as their occupancy and color of the imaged scene at the specified 3D location. 
NeRF~\cite{mildenhall2020nerf} models 3D scenes by mapping both 3D coordinates and the viewing direction to RGBA space, achieving breakthrough performances in novel view synthesis. 
Multiple works have extend the NeRF paradigm targeting higher quality reconstruction~\cite{barron2021mip, zhang2020nerf++} and faster runtime~\cite{liu2020neural, yu2021plenoxels, yu2021plenoctrees}.
Two directions, particularly related to object pose estimation, is the extension of NeRF beyond the strict single-scene setting and the removal on the dependency for training time poses.

Growing out of the fixed-scene setting, NeRF in the Wild~\cite{martin2021nerf} learns to aggregate views of the same scene taken in different settings by learning image-specific embeddings, while Nerfies~\cite{park2021nerfies} learn to deform rays to represent deformable objects. 
PixelNeRF~\cite{yu2021pixelnerf} learns scene-based dense embeddings to represent multiple scenes.
CodeNeRF~\cite{jang2021codenerf} disentangles shape and texture across instances from the same category. 
These methods, however, require ground-truth camera poses during training. 
Though several generative methods~\cite{schwarz2020graf, niemeyer2021giraffe, gu2021stylenerf} have been proposed to model object categories, they are unable to estimate poses, and employ neural rendering that can violate the scene geometry as in SSV~\cite{mustikovela2020self}. 
Unlike them, through the use of an implicit 3D representation and analytical rendering, our reconstructions are 3D consistent.

Multiple pose-free NeRF methods~\cite{wang2021nerf--, lin2021barf, jeong2021self-cal, chng2022garf} have been proposed that attempt to learn camera poses during training by refining initial pose estimates through gradients coming from the NeRF model itself.
However, these methods are only pose-free on forward-facing scenes, needing COLMAP as initialization for 360\degree scenes, if they even are evaluated on such challenging settings. 
While NeRF methods can be used to retrieve the pose of new images under certain condition~\cite{yen2020inerf, jang2021codenerf}, they require expensive test-time optimization.
By comparing image reconstructions based on initial noisy pose estimates to the target image, they perform many  gradient descent steps on the camera parameters to gradually align the two images. 
In addition to being a slow process, this approach can get trapped in local minima in the multi-object setting. 
In comparison, our model can predict the pose of unseen instances in a single forward pass.
A notable exception from other pose-free NeRFs is GNeRF~\cite{meng2021gnerf}, which can operate without initialization thanks to its adversarial training. 
However, it is limited to single scenes and is slow to train as a result of the additional GAN-based objective. 
A comparison to related NeRF-based approaches is shown in~\cref{tab:models_table}.

\section{Method}
\label{sec:method}
\begin{figure}[t]
    \centering
    \includegraphics[width=\linewidth]{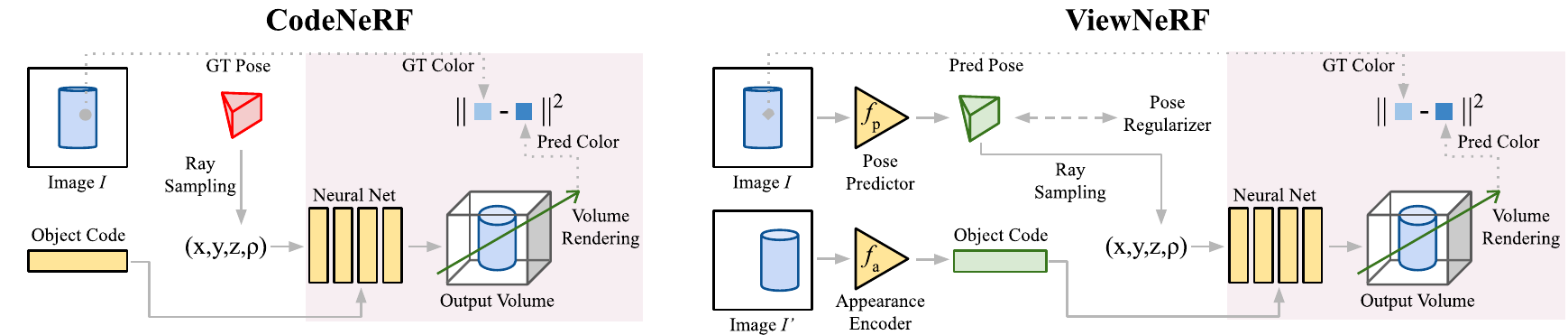}
    \vspace{5pt}
    \caption{(Left) CodeNeRF~\cite{jang2021codenerf} requires ground-truth pose information at training time and performs expensive direct optimization for each object appearance code. 
    In practice, CodeNeRF also enforces that the object code for distinct views of the same object are the same which provides some multi-view signal. 
    (Right) In contrast, our ViewNeRF approach is fully self-supervised by making use of a separate pose predictor $f_p$ and appearance encoder $f_a$ that can be applied to any image in a single-shot fashion. 
    \vspace{-10pt}
    }
    \label{fig:method_overview}
\end{figure}

Here we outline the main components and training procedure for our ViewNeRF model. 
Unlike prior methods, ViewNeRF is capable of estimating the pose for held out images and can be trained on images of multiple instances (\ie from different scenes) of the same object category \emph{without} requiring any ground-truth pose supervision. 

Our goal is to estimate the viewpoint (\ie camera pose) of an unseen object instance from a known category (\eg car, chair, \etc) in an image.
To this end, we wish to learn a function, $f_p$ that takes an image $I$ as input and outputs the corresponding viewpoint represented as the rotation and translation, \ie $p=f_p(I)$.
As ground-truth viewpoints are not available for training, we treat learning pose prediction as an image reconstruction problem. 
In the same vein as \cite{tulsiani2018multi, insafutdinov2018unsupervised, yu2021pixelnerf, jang2021codenerf, mariotti2021viewnet}, we exploit multi-view information in the form of image pairs. 
Specifically, given $N$ unlabeled image pairs $\{I_n,I'_n\}_{n=1}^N$, where each pair contains source and target images of the same object instance which differ only in their unknown viewpoints, our objective is to reconstruct the target image $I$ from the source image $I'$.

Clearly, one needs a good estimate of the viewpoint of $I$ in order to reconstruct it from $I'$.
During training we reconstruct the target image $I$ using its estimated viewpoint $f_p(I)$ and the appearance information extracted from our viewpoint-independent appearance encoder $f_a$ for the source image $I'$. 
For rendering, we pass the pose and appearance information to a NeRF-based decoder $f_r$, to reconstruct the target image $I$, and minimize an image reconstruction loss to simultaneously learn the weights of $f_p$, $f_a$, and $f_r$ which are instantiated as neural networks,
\begin{equation}
    \label{eq:recon_error}
    \min_{f_p,f_a,f_r}\frac{1}{N} \sum_{n=1}^{N} \mathcal{L}(f_r(f_a(I'_n), f_p(I_n)),I_n)+\lambda \mathcal{L}_{\text{reg}}(f_p(I_n)).
\end{equation} $\mathcal{L}$ is a loss function that measures the difference between the reconstructed and target images, and $\mathcal{L}_{\text{reg}}$ is a pose regularization term applied to the viewpoint predictions, which is weighted by a scalar $\lambda$. 
The training pipeline for our model, ViewNeRF, is shown in \cref{fig:method_overview} (right).

Clearly the target image cannot be successfully reconstructed without its viewpoint information.
However, in the case of an arbitrary decoder, where the appearance and viewpoint encodings are passed through multiple arbitrary nonlinear transformations, there are at least two challenges: 
it is not guaranteed that (i) the estimated viewpoints are disentangled from the appearance and, if they are, (ii) the estimated viewpoints are geometrically meaningful transformations.
Hence, it is crucial that the decoder utilizes the estimated viewpoint in a geometrically consistent way.

\vspace{-6pt}
\subsection{NeRF decoder - $f_r$}
\vspace{-3pt}
NeRFs are originally formulated as 3D models that learn to reconstruct what a scene looks like when observed from a specified viewpoint $p$. 
Formally, they combine a volume rendering operation~\cite{kajiya1984ray} with a neural network that learns to map an input $(\mathbf{x},\rho) \in \mathrm{R}^3 \times \mathcal{S}^2$, consisting of a 3D coordinate and a viewing direction, to a 4D vector $(r,g,b,\sigma)$ in RGBA space.
To reconstruct an image, rays corresponding to each pixel are cast from a camera $p$, and the network is queried multiple times along each ray to estimate the color and occupancy at those locations. 
Then, values along the ray are integrated according to their density to form a single pixel value. An important property of NeRF is that the predicted occupancy $\sigma$ only depends on the input coordinates $\mathbf{x}$. 
This coupled with its analytical rendering, grants them relatively strong 3D consistency (although not perfect~\cite{zhang2020nerf++}). Along with its high resolution, this property makes it particularly suitable as a decoder in our pose estimation pipeline.

\noindent{\bf Object appearance conditioned on a NeRF decoder.}
Standard NeRFs are trained to model a single 3D scene, effectively memorizing its shape and appearance from multiple viewpoints. 
In a category-based setting, this would mean training an individual model for each object instance, which would be very time consuming and no information across instances would be shared. 
Hence, a better approach is to implement a conditioning mechanism to allow the NeRF decoder $f_r$ to reconstruct specific instances. 
This conditioning should be pose-agnostic in order to let the pose predictor $f_p$ learn to disentangle pose from appearance. 
Therefore, we adopt a strategy inspired by ViewNet~\cite{mariotti2021viewnet} and CodeNeRF~\cite{jang2021codenerf} where object instances are fully described by a latent object code $\mathbf{a}$. 
Similar to~\cite{jang2021codenerf}, $\mathbf{a}$ is mapped at different depths of the NeRF model to condition its activations, and similar to~\cite{mariotti2021viewnet}, $\mathbf{a}$ is predicted by an appearance network $f_a$ that learns a global latent space shared over all object instances.

\vspace{-6pt}
\subsection{Pose estimator - $f_p$}
\vspace{-3pt}
Conventional NeRF methods requires ground-truth camera poses during training. 
Recent extensions~\cite{yen2020inerf, wang2021nerf--, jeong2021self-cal, lin2021barf, chng2022garf} allows for estimating camera poses with NeRF by letting reconstruction gradients flow to the camera parameters.
However, this approach suffers from two issues: i) it is computationally expensive, requiring hundreds of steps to converge, if at all, and ii) it can get stuck in local minima, limiting its operation to forward-facing scenes when a reasonable pose initialization is not available. 
Following the recent advances in viewpoint estimation~\cite{tulsiani2018multi, insafutdinov2018unsupervised, mustikovela2020self, mariotti2021viewnet}, we posit that a better solution is to estimate poses directly from images using a pose predictor $f_p$. 
This enables fast prediction during inference and generalizes efficiently to new object instances.
However, $f_p$ is still subject to local minima and may not receive meaningful gradients from $f_r$, as reconstruction errors can arise either from the reconstruction process or an incorrect pose prediction.
This can lead to the collapse of pose predictions and to degenerate solutions where the model relies only on the appearance encoding $\mathbf{a}$ to reconstruct $I'$. 
Next we introduce two mechanisms to prevent this.

\noindent{\bf Multi-hypothesis predictions.} We supply $f_p$ with a multi-head predictor as used in~\cite{insafutdinov2018unsupervised,mariotti2021viewnet}.
During training, we let the pose estimator output multiple hypotheses $f_p(I) =  p_1, \ldots, p_K$, and each of them is fed to $f_r$ to produce a low resolution reconstruction. 
These are then compared to the target, and the pose $p^*$ resulting in the best reconstruction is selected. 
$p^*$ is then passed again to the NeRF decoder, this time at full resolution. 
For inference, a student head is jointly trained to predict $p^*$, removing the need for multiple outputs~\cite{mariotti2021viewnet}. 

\noindent{\bf Pose regularization.}
Multiple pose predictions alone are not always sufficient to prevent training collapse or instability as all heads can still predict the same pose. 
Inspired by generative models like~\cite{niemeyer2021giraffe, gu2021stylenerf} that sample poses during training, we encourage the predicted pose distribution to follow a  prior distribution $\mathcal{P}$. 
As generative models do not aim to reconstruct images from a specific viewpoint, they can directly sample a pose from the prior $p \sim \mathcal{P}$ and use it to generate an image. 
However, this is unfeasible in our case, as a random pose would not match that of the specific image that we would like to reconstruct. 
Instead, we attempt to match batch-wise distributions, following the assumption that a batch of predicted poses should closely follow the pose prior $\mathcal{P}$.

Specifically, given a batch of $B$ predicted poses $p^*_{1, \ldots, B}$, we sample $K$ pseudo-targets $p'_{1, \ldots, K} \sim \mathcal{P}$ and compute for each $p'_i$ its closest match $p^*_j$ in the batch. Finally, the distance ${||p'_i - p^*_j||^2}$ is added to the loss, \ie $\mathcal{L}_{\text{reg}} = \frac{1}{K} \sum_{i=1}^K \min_{j=1\ldots B} ||p'_i - p^*_j||^2$.
This prevents the collapse of all predictions to a single point while being very cost-efficient. To prevent unnecessary noise, the regularization weight $\lambda$ in \cref{eq:recon_error} is progressively tuned down during training. 
Further details are provided in the supplementary.

\vspace{-6pt}
\subsection{Reconstruction objective}
\vspace{-3pt}
Reconstructing full resolution images requires millions of queries to the NeRF decoder and hence is expensive, so NeRFs are usually trained by sampling a subset of pixels per image per iteration.
This strategy works well when using ground-truth camera poses, as each sampled pixel corresponds to one exact ray that will stay constant during training. 
However, when jointly estimating poses and training the NeRF model, it can introduce a significant amount of noise, as the randomly selected pixels might not contain enough relevant information to recover incorrectly estimated poses. In particular, some object categories such as cars can exhibit symmetries that can only be broken by focusing on fine  details (\eg color of the headlights). 

To this end, we deviate from the standard NeRF training procedure and instead use reconstructions of the entire image, however, in low-resolution coupled with a perceptual loss~\cite{johnson2016perceptual} (denoted as $\mathcal{L}$ in \cref{eq:recon_error}) that provides a finer structure-preserving objective. 
As the perceptual loss aims to match activation maps from a pretrained network, it is more sensitive to salient image features like edges that would be missing or misplaced under wrongly estimated poses.

\section{Experiments}
\label{sec:experiments}
\subsection{Implementation details}

We use EfficientNet~\cite{tan2019efficientnet} backbones for our pose and appearance encoders $f_p$ and $f_a$, and a compact NeRF model with two fully connected layers with 128 dimensions for the decoder.
Note that our decoder is designed to be significantly smaller than the one used in~\cite{meng2021gnerf} and~\cite{jang2021codenerf} to discourage unnecessarily complex mappings and to make it more efficient. 
Computational efficiency is particularly important in our experiments, as we use a perceptual loss for reconstruction error which requires generating a full image. 

\noindent{\bf Camera pose prediction.}
To link the predictions of $f_p$ to real world poses, we need to ensure it can be interpreted as such. Similar to ViewNet and GNeRF, we formulate pose as a point on the 3D unit sphere $\mathcal{S}^2$ from which we derive a camera matrix using a Gram-Schmidt orthogonalization process. While other representations like quaternions are possible, this provides a simple way to enforce the necessary constraints and exhibits favorable properties for optimization~\cite{zhou2019continuity}.
For synthetic datasets, we follow GNeRF's assumption that the object is located at the center of the scene where the camera is pointed to, and that the camera is held upright, \ie it is aligned with the $z$ vector in world coordinates. On the Freiburg Cars dataset~\cite{sedaghat2015unsupervised}, this assumption is not valid as data was hand-recorded. Therefore, we additionally allow our viewpoint estimator to predict a camera distance, target point, \ie a point along the camera principal axis, and an upwards direction to account for in-plane rotation. Instead of hard constraints, we set soft target for the first 10 epochs of training.
Additional implementation details can be found in the supplementary material.

As poses are predicted up to an arbitrary rotation, following the evaluation in~\cite{mariotti2021viewnet}, we align our estimated camera poses with the ground-truth labels by solving an orthogonal Procrustes problem.
As our main goal is to estimate viewpoint from single images rather than high-fidelity reconstruction, we evaluate our method in terms of viewpoint accuracy, reporting rotation and translation errors. 
The other approaches we compare to, \cite{meng2021gnerf, jang2021codenerf, mariotti2021viewnet}, each use their own sets of metrics making direct comparison difficult. 
Hence, for fair comparison, we re-evaluate their models and report viewpoint accuracy with a 10\degree~threshold, along with median rotation and translation error for each method in all experiments.
Taking the median instead of the average provides a less noisy estimate in the presence of strong symmetries~\cite{tulsiani2015viewpoints}. 
Translations errors are normalized by the camera distance to the origin to account for scaling differences.

\vspace{-5pt}
\subsection{Multi-instance results}

\begin{figure}[t]
    \centering
    \resizebox{\linewidth}{!}
    {
    \setlength{\tabcolsep}{05pt}
    \begin{tabular}{ccccc}
          {\small \shortstack{CodeNeRF\\w/o init}} & {\small \shortstack{CodeNeRF\\w/ init}} & {\small \shortstack{ViewNet\\{\color{white}I}}}  & {\small \shortstack{ViewNeRF\\ (Ours)}} & {\small \shortstack{Ground\\Truth}}\\
          
             \includegraphics[width=.15\textwidth]{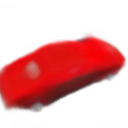}
        &    \includegraphics[width=.15\textwidth]{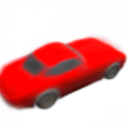}
        &    \includegraphics[width=.15\textwidth]{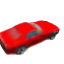}
        &    \includegraphics[width=.15\textwidth]{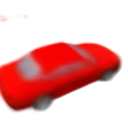}
        &    \includegraphics[width=.15\textwidth]{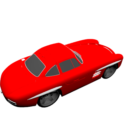}\vspace{-3pt}\\
        
             \includegraphics[width=.15\textwidth]{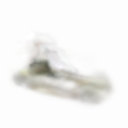}
        &    \includegraphics[width=.15\textwidth]{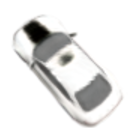}
        &    \includegraphics[width=.15\textwidth]{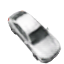}
        &    \includegraphics[width=.15\textwidth]{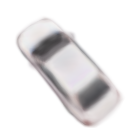}
        &    \includegraphics[width=.15\textwidth]{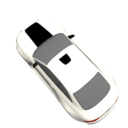}\vspace{-3pt}\\

             \includegraphics[width=.15\textwidth]{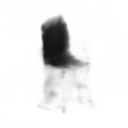}
        &    \includegraphics[width=.15\textwidth]{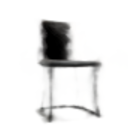}
        &    \includegraphics[width=.15\textwidth]{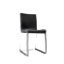}
        &    \includegraphics[width=.15\textwidth]{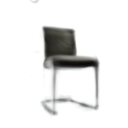}
        &    \includegraphics[width=.15\textwidth]{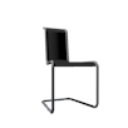}\vspace{-3pt}\\
            
             \includegraphics[width=.15\textwidth]{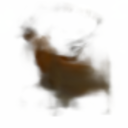}
        &    \includegraphics[width=.15\textwidth]{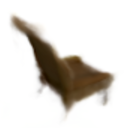}
        &    \includegraphics[width=.15\textwidth]{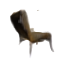}
        &    \includegraphics[width=.15\textwidth]{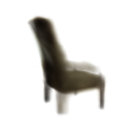}
        &    \includegraphics[width=.15\textwidth]{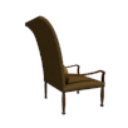}\vspace{-3pt}
    \end{tabular}
    }
    \vspace*{10pt}
    \caption{Comparison of reconstructions from the supervised CodeNeRF, unsupervised ViewNet, and our unsupervised ViewNeRF.
    \vspace*{-15pt}
    }
    \label{fig:ShapeNet_visualizations}
\end{figure}

\begin{table}[t]
    \centering
    \renewcommand{\b}{\bfseries}
    \resizebox{1.0\linewidth}{!}
    {
    \begin{tabular}{l|cc|cc|cc|cc|}
                                                       & \multicolumn{4}{c|}{Supervised} & \multicolumn{4}{c|}{Unsupervised} \\
                                                       & \multicolumn{2}{c}{CodeNeRF, w/o init} & \multicolumn{2}{c|}{CodeNeRF, w/ init} & \multicolumn{2}{c}{ViewNet} & \multicolumn{2}{c|}{Ours} \\ 
                                                       & car  & chair & car  & chair & car  & chair & car  & chair \\
                                                       \hline
         Accuracy at 10\degree (\%, $\uparrow$)        & 08.5 & 03.4  & \b 82.1 & \b 60.2  & 61.2 & 76.7  & \b 70.0 & \b 82.8  \\
         Median rotation error (\degree, $\downarrow$) & 115  & 108   & \b 3.53 & \b 7.70  & 6.54 & 4.25  & \b 5.71 & \b 4.18  \\
         Median translation error (\%, $\downarrow$)   & 139  & 134   & \b 5.9  & \b 13.9  & n/a  & n/a   & \b 8.0  & \b 6.4  
    \end{tabular}
    }
    \vspace{15pt}
    \caption{Multi-instance results on ShapeNet-SRN. 
    CodeNeRF pretrained models were kindly provided by the authors. When initialized, pose estimates were randomly drawn within 30\degree of the ground-truth, where \textbf{bold} results indicate the best model per category.}
    \label{tab:Shapenet_res}
\end{table}

In \cref{tab:Shapenet_res} we first evaluate our ViewNeRF approach on the ShapeNet-SRN dataset~\cite{sitzmann2019scene} which contains renderings of ShapeNet~\cite{chang2015shapenet} cars and chairs. 
We compare to CodeNeRF~\cite{jang2021codenerf}, a \emph{supervised} NeRF-based model and the \emph{unsupervised} voxel-based ViewNet~\cite{mariotti2021viewnet}. 
While ViewNet reports results on ShapeNet, it is only trained on a limited set of viewpoints, \ie the elevation of views only spans $[-20\degree, 40\degree]$, instead of the full range in ShapeNet-SRN. Hence, we retrain it using code from the authors on this new data split.

CodeNeRF requires expensive test-time optimization to perform pose estimation and only reports results for a \emph{single} object instance in their paper, \ie not multiple instances from the same category, starting from hand-selected poses\footnote{Confirmed via correspondence with the authors.}. 
Therefore, we re-evaluated it on each test instance, under two settings, a realistic one where the starting pose is uniformly sampled according to the training distribution (`w/o init'), and an easier setting, in which the initial pose is chosen to be within 30\degree~of the ground-truth (`w/ init').

The results in \cref{tab:Shapenet_res} illustrate that CodeNeRF, despite being trained with ground-truth pose, is unable to properly estimate pose when the initial estimate is noisy. Since both pose and object embeddings have to be jointly optimized, the process can converge to a degenerate solution, relying mostly on the appearance embeddings rather than pose to minimize the reconstruction error. 
Test-time reconstructions shown in \cref{fig:ShapeNet_visualizations} confirm this. 
While reconstruction with good initializations are accurate, noisy initialization results in poor reconstructions. 
Finally, compared with ViewNet, our approach reaches higher pose prediction performance by making fewer gross pose errors, \eg ViewNet predicts in the wrong orientation for the car in the second row of \cref{fig:ShapeNet_visualizations}.

\subsection{Real scenes results}

\begin{figure}[t]
    \centering
    \resizebox{\linewidth}{!}
    {
    \includegraphics{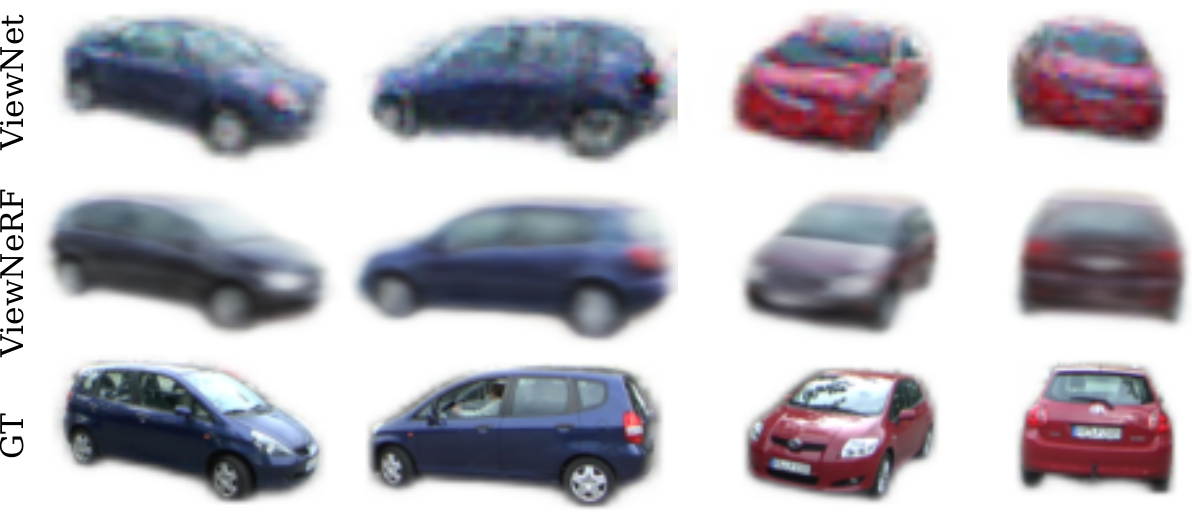}
    }
    \vspace{-5pt}
    \caption{Comparison of reconstructions from the unsupervised ViewNet and ViewNeRF on held-out test instances from the Freiburg Cars dataset.
    }
    \label{fig:Freiburg_visualizations_compact}
\end{figure}

Here we demonstrate ViewNeRF's ability to work on real images using the Freiburg Cars dataset~\cite{sedaghat2015unsupervised}. 
The target car instance in each image is first segmented using MaskRCNN~\cite{he2017mask}, and out of the 48 scenes, the first 40 are used for training, the next three for validation, and the remaining five for testing. As the data is only labeled with weak viewing direction information, we only report rotation-base metrics.
The results in \cref{tab:FCar_res} illustrate a large gap between the performances of our approach and ViewNet. 
We mostly attribute it to ViewNet's inability to model the complex illumination patterns (\eg reflections) on real cars.
Qualitative results in \cref{fig:Freiburg_visualizations_compact} further illustrate this, \ie reconstructions from ViewNet, while having sharper colors are very noisy. In addition, it seems that ViewNet is unable to differentiate the front from the back of the red car.

\begin{table}[t]
    \centering
    \renewcommand{\b}{\bfseries}
    \resizebox{\linewidth}{!}
    {
    \begin{tabular}{l|cc|ccccc}
        & ViewNet   & Ours    &  \shortstack{Ours,\\no predictor} &  \shortstack{Ours,\\MSE} &  \shortstack{Ours,\\singlehead} &  \shortstack{Ours,\\subsampled} &  \shortstack{Ours,\\no reg.} \\ \hline
         Accuracy at 10\degree (\%, $\uparrow$)            & 50.0      & \b 73.5 & 00.8 & 04.1  & 11.4 & 13.5 & 54.5\\
         Median rotation error (\degree, $\downarrow$)    & 9.99      & \b 8.05 & 90.8 & 91.0  & 67.4 & 93.1 & 9.09\\
    \end{tabular}
    }
    \vspace{10pt}
    \caption{Comparison to ViewNet and ablated versions of our ViewNeRF method on the Freiburg Cars dataset.}
    \label{tab:FCar_res}
    \vspace{-10pt}
\end{table}

\noindent{\bf Ablated models.}
To validate our design choices, we also evaluate multiple ablated versions of our model on the Freiburg Car dataset. We evaluate four variations: (i) removing estimators and trying to learn poses directly with backpropagation, \ie standard pose-free NeRF training, (ii) using a MSE loss instead of the perceptual loss for reconstruction, (iii) using a single pose hypothesis, (iv) subsampling 1024 random pixels per image instead of using low resolution full reconstructions, and (v) removing pose regularization $\mathcal{L}_\text{reg}$. 
All ablations produce worse performance, with the first three resulting in catastrophic failure.

\subsection{Single instance results}
\begin{table}[t]
    \centering
    \renewcommand{\b}{\bfseries}
    \resizebox{0.85\linewidth}{!}
    {
    \begin{tabular}{c|ccc|ccc}
         & \multicolumn{3}{c|}{GNeRF} & \multicolumn{3}{c}{Ours}\\
         & Acc (\%,$\uparrow$) & MR (\degree,$\downarrow$) & MT (\%,$\downarrow$) & Acc (\%,$\uparrow$) & MR (\degree,$\downarrow$) & MT (\%,$\downarrow$)  \\ \toprule
         Chair  & \b 100   & \b 2.645 & \b 4.401 & \b 100   &    3.012 &    4.680 \\
         Drums  & \b 98.5  & \b 3.307 & \b 5.489 &    80.5  &    5.212 &    8.356 \\
         Hotdog &    74.5  &    7.120 &    11.12 & \b 96.0  & \b 2.412 & \b 3.898 \\
         Lego   & \b 91.5  &    5.153 &    8.313 &    87.0  & \b 4.659 & \b 7.571 \\
         Mic    & \b 97.5  & \b 3.022 & \b 4.787 &    93.5  &    4.169 &    6.823 \\
         Ship   &    15.0  &    28.23 &    43.56 & \b 69.5  & \b 6.674 & \b 9.946 \\
    \end{tabular}
    }
    \vspace{10pt}
    \caption{NERF synthetic single scenes. Acc: Accuracy at 10\degree, MR: Median rotation error, MT: Median translation error. GNeRF models were retrained using published code.
    }
    \label{tab:single_scene}
    \vspace{-10pt}
\end{table}

Finally we evaluate ViewNeRF in the single-scene setting with full 360\degree~ rotations on the synthetic-NeRF datasets~\cite{mildenhall2020nerf} used in GNeRF~\cite{meng2021gnerf}. 
GNeRF is closely related to our model as it can estimate pose with a simple single forward pass. 
However, the pose results reported in the original GNeRF paper are from the training split of the data, where they are learned using a mixture of gradient descent optimization and soft-labeling. Similarly, COLMAP~\cite{schonberger2016structure} is evaluated directly over the training images.
In \cref{tab:single_scene} we instead evaluate the GNeRF pose predictor on the test split in order to perform a fair comparison with our model. 
We observe that in spite of its strong performance on the training split and the much larger model it uses, GNeRF results are broadly comparable to ours during inference.
This can be explained by the size of the training set, that only contains 100 samples, hinting towards overfitting. 
The ship scene exhibits a strong rotational symmetry, and is thus  particularly challenging for both methods.

\vspace{-5pt}
\subsection{Limitations}
Although our model outperforms prior works in category-based viewpoint estimation, it also has certain limitations that hinders its applicability on more complex scenarios.
While the requirement for multiple views at training time is common, it limits our approach to  multi-view datasets. 
It would be desirable to have a model that can learn object categories from different instances without requiring multi-view data. 
While generative methods (\eg~\cite{nguyen2019hologan, mustikovela2020self, niemeyer2021giraffe, gu2021stylenerf}) possibly possess this ability, they also employ neural-based decoders that hurt 3D consistency.
Another limitation is the need for segmenting foreground object from cluttered background. 
We observed that when using unsegmented views, complex backgrounds that form the majority of an image prevented NeRF from paying enough attention to the object to capture the details needed for estimating category-level pose.
Forcing our model to focus less on the background during training by using two separate NeRFs as in~\cite{niemeyer2021giraffe} could be a potential solution. 

\section{Conclusion}
\label{sec:conclusion}
We presented ViewNeRF, a conditional NeRF-based method for accurate category-centric pose estimation that is trained from self-supervision alone. 
Through careful design choices, our ViewNeRF model manages to predict accurate viewpoints during training and testing across a wide variety of real and synthetic datasets, going beyond what previous NeRF-based models are capable of. 
Through extensive evaluation, we show the pitfalls of the gradient descent-based pose recovery that is used in many NeRF pipelines. 
We compare our method with other related self-supervised approaches and illustrate the benefits of NeRF over explicit 3D modeling for the challenging task of single image pose estimation.

\vspace{5pt}
\noindent\textbf{Acknowledgment.} HB is supported by the EPSRC Visual AI grant EP/T028572/1.

\clearpage
\bibliography{bibliography}

\newpage
\appendix
\section{Pose regularization}

We provide pseudo-code for our pose regularization method. Note that $K$ might not be equal to $B$. In practice, instead of using the minimal distance, we use a soft minimum to decrease noise.
The pose prior approximately follows the training data distribution, i.e. top half of the sphere on NeRF scenes, uniform on ShapeNet, ground level for Freiburg cars. 
Regularization strength $\lambda$ starts at 1 and undergoes exponential scheduling, being multiplied by 0.1 every 10 epochs before being turned off at epoch 30.

\begin{algorithm}[h]
 \KwInput{Minibatch of predicted poses $p^*_{1,\ldots,B}$, Prior distribution $\mathcal{P}$, number of samples $K$}
 \KwOutput{Regularization loss $\mathcal{L}_{reg}$}
 $\mathcal{L}_{reg} = 0$
 
 \For{$i \in 1\ldots K$}{
    $p' \sim \mathcal{P}$ \tcp*{draw a pseudo-target from $\mathcal{P}$}
    $\text{dists} =  ||p^* - p'||$ \tcp*{distance between each predicted pose and p', size $B$}
    $\text{weights} = SoftMax(-\text{dists})$ \tcp*{Batch-wise SoftMax}
    
    $\text{weighted\_dists} = \text{weights} * \text{dists}$
    
    $\mathcal{L}_{reg} += \frac{1}{K} * Avg(\text{weighted\_dists})$ \tcp*{Batch-wise Average}
 }
 \caption{Pose regularization}
\end{algorithm}

\section{Implementation details}

\noindent{\bf Perceptual loss.}
We implement perceptual loss using a pretrained VGG16 model. The total loss consists of standard MSE in pixel space, plus MSE between features extracted before the first, second and third max polling layers of the model, with weights 10, 1, 1, and 1 respectively. Because of the cost of producing full images, the output reconstruction uses a 64x64 resolution.

\begin{figure}[t]
    \centering
    \includegraphics[width=.75\textwidth]{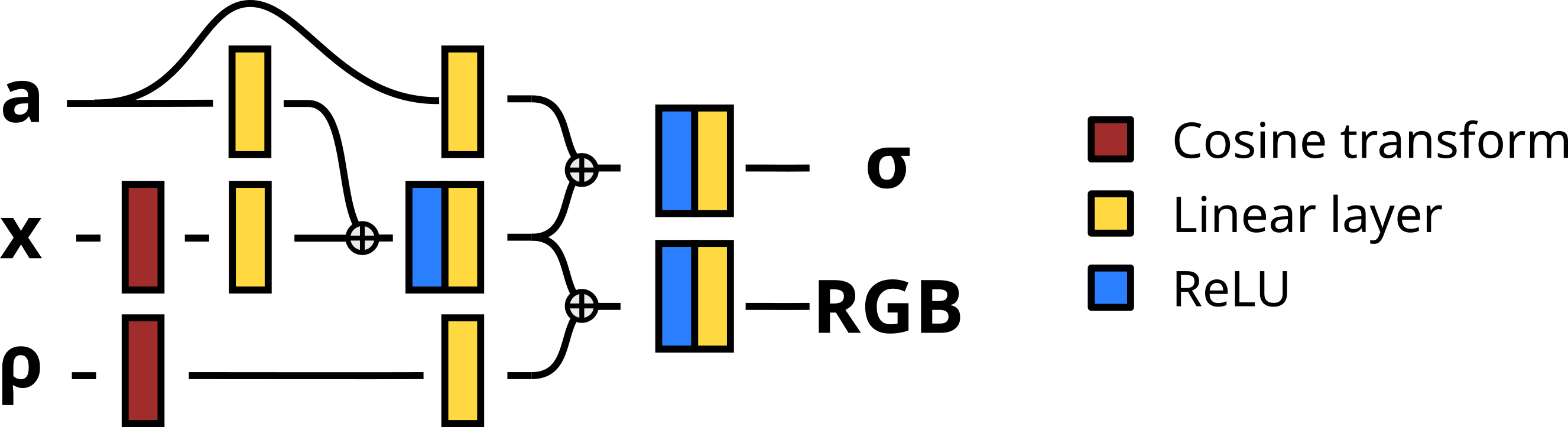}
    \vspace{15pt}
    \caption{NeRF archtitecture used in ViewNeRF. \textbf{a} depicts the appearance embedding, while \textbf{x} and \textbf{\textrho} are the spatial coordinate and viewing direction.}
    \label{fig:arch}
\end{figure}

\noindent{\bf NeRF architecture.}
The architecture of our NeRF decoder is depicted in \cref{fig:arch}. To encourage 3D consistency, we use cosine embeddings of size 8 and 1 for \textbf{x} and \textbf{\textrho} respectively. They are then mapped with linear layers to the inner dimension of the model which is 128.

\section{Supplemental results}
\subsection{Reconstruction metrics}
\vspace{-5pt}

While ViewNeRF is not designed for accurate reconstructions, quantitative values could be useful for future references. SSIM and PSNR for ShapeNet are shown in \cref{tab:recons_metrics}.

\begin{table}[t]
    \vspace{-20pt}
    \begin{subtable}{.45\linewidth}
        \resizebox{\linewidth}{!}
        {
        \begin{tabular}{l|rrr}
            & \shortstack{ShapeNet\\cars} & \shortstack{ShapeNet\\chairs} & \shortstack{Freiburg\\cars}\\\midrule
             PSNR ($\uparrow$)&  15.6 & 17.5 & 19.1\\
             SSIM ($\uparrow$)&  0.70 & 0.77 & 0.84
        \end{tabular}
        }
        \caption{ViewNeRF reconstruction metrics on evaluation instances.}
        \label{tab:recons_metrics}
    \end{subtable}
    \hfill
    \begin{subtable}{.5\linewidth}
    
        \vspace{20pt}
        \resizebox{\linewidth}{!}
        {
        \begin{tabular}{l|rrr}
                      & CodeNeRF & ViewNet & ViewNeRF \\
                      \midrule
            Train     & $139\pm9.6$ & $9.1\pm 3.7$ & $14.7\pm 1.7$\\
            Inference & $38.4\pm0.6\times 10^3$& $0.1\pm 0.3$ &  $0.4\pm 1.0$\\
        \end{tabular}
        }
    \caption{Processing time per sample in milliseconds (ms), reported as mean and standard deviation averaged over 100 batches.}
    \label{tab:processsing_time}
    \end{subtable}
\end{table}

\subsection{Extra visualizations}
\vspace{-5pt}

\begin{figure}[t]
    \centering
    \setlength{\tabcolsep}{0pt}
    {
    \begin{tabular}{cccccc}
            ViewNet & ViewNeRF & ViewNet & ViewNeRF & ViewNet & ViewNeRF \\
          
              \includegraphics[width=.15\textwidth]{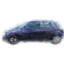}
            & \includegraphics[width=.15\textwidth]{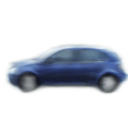}
            & \includegraphics[width=.15\textwidth]{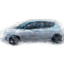}
            & \includegraphics[width=.15\textwidth]{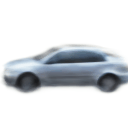}
            & \includegraphics[width=.15\textwidth]{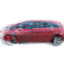}
            & \includegraphics[width=.15\textwidth]{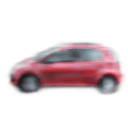}\\
            
              \includegraphics[width=.15\textwidth]{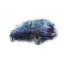}
            & \includegraphics[width=.15\textwidth]{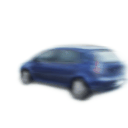}
            & \includegraphics[width=.15\textwidth]{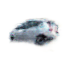}
            & \includegraphics[width=.15\textwidth]{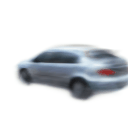}
            & \includegraphics[width=.15\textwidth]{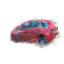}
            & \includegraphics[width=.15\textwidth]{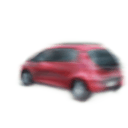}\\
            
              \includegraphics[width=.15\textwidth]{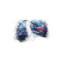}
            & \includegraphics[width=.15\textwidth]{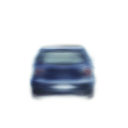}
            & \includegraphics[width=.15\textwidth]{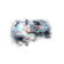}
            & \includegraphics[width=.15\textwidth]{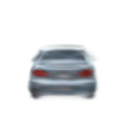}
            & \includegraphics[width=.15\textwidth]{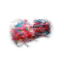}
            & \includegraphics[width=.15\textwidth]{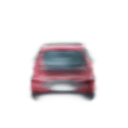}\\
            
             \includegraphics[width=.15\textwidth]{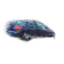}
            & \includegraphics[width=.15\textwidth]{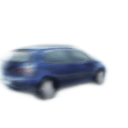}
            & \includegraphics[width=.15\textwidth]{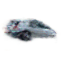}
            & \includegraphics[width=.15\textwidth]{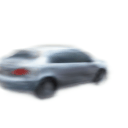}
            & \includegraphics[width=.15\textwidth]{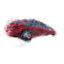}
            & \includegraphics[width=.15\textwidth]{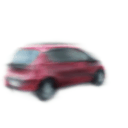}\\
              
              \includegraphics[width=.15\textwidth]{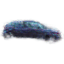}
            & \includegraphics[width=.15\textwidth]{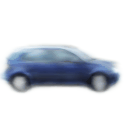}
            & \includegraphics[width=.15\textwidth]{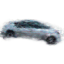}
            & \includegraphics[width=.15\textwidth]{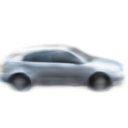}
            & \includegraphics[width=.15\textwidth]{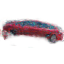}
            & \includegraphics[width=.15\textwidth]{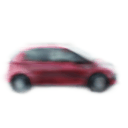}\\
              
              \includegraphics[width=.15\textwidth]{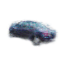}
            & \includegraphics[width=.15\textwidth]{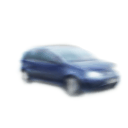}
            & \includegraphics[width=.15\textwidth]{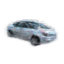}
            & \includegraphics[width=.15\textwidth]{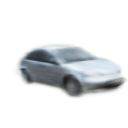}
            & \includegraphics[width=.15\textwidth]{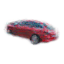}
            & \includegraphics[width=.15\textwidth]{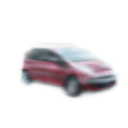}\\
              
             \includegraphics[width=.15\textwidth]{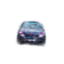}
            & \includegraphics[width=.15\textwidth]{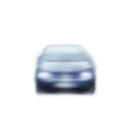}
            & \includegraphics[width=.15\textwidth]{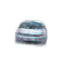}
            & \includegraphics[width=.15\textwidth]{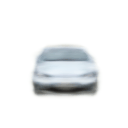}
            & \includegraphics[width=.15\textwidth]{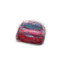}
            & \includegraphics[width=.15\textwidth]{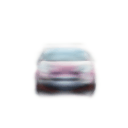}\\
            
              \includegraphics[width=.15\textwidth]{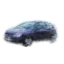}
            & \includegraphics[width=.15\textwidth]{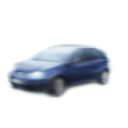}
            & \includegraphics[width=.15\textwidth]{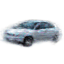}
            & \includegraphics[width=.15\textwidth]{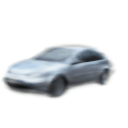}
            & \includegraphics[width=.15\textwidth]{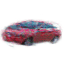}
            & \includegraphics[width=.15\textwidth]{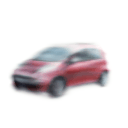}\\

              \multicolumn{2}{c}{\includegraphics[width=.15\textwidth]{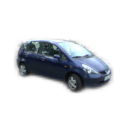}}
            & \multicolumn{2}{c}{\includegraphics[width=.15\textwidth]{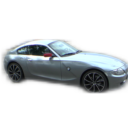}}
            & \multicolumn{2}{c}{\includegraphics[width=.15\textwidth]{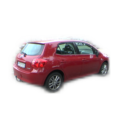}}\\
    \end{tabular}
    }
    \caption{Reconstructions of Freiburg car test instances for ViewNet and ViewNeRF. Bottom row is the frame used for providing appearance embedding.}
    \label{fig:FC_recons}
\end{figure}

In \cref{fig:FC_recons}, we provide extra comparison between our method and ViewNet on Freiburg cars, by sampling views at a 45\degree~interval around reconstructed test instances. It is apparent that ViewNet does not manage to reconstruct the back of the car correctly.

\section{Speed analysis}
Table \ref{tab:processsing_time} shows the time taken to process 1 64x64 image on a Tesla V100. The inference time of CodeNeRF, requiring 300 gradient descent steps, is 5 orders of magnitude higher than ours, while its larger NeRF backbone also makes it more expensive during training.

\end{document}